# Innovative Adaptive Imaged Based Visual Servoing Control of 6 DoFs Industrial Robot Manipulators

**Rongfei Li, University of California, Davis, California, Email: rfli@ucdavis.edu.**

**Francis F. Assadian, University of California, Davis, California 95618, Email: fassadian@ucdavis.edu.**

## Abstract

Image-based visual servoing (IBVS) methods have been well developed and used in many applications, especially in pose (position and orientation) alignment. However, most research papers focused on developing control solutions when 3D point features can be detected inside the field of view. This work proposes an innovative feedforward-feedback adaptive control algorithm structure with the Youla Parameterization method. A designed feature estimation loop ensures stable and fast motion control when point features are outside the field of view. As 3D point features move inside the field of view, the IBVS feedback loop preserves the precision of the pose at the end of the control period. Also, an adaptive controller is developed in the feedback loop to stabilize the system in the entire range of operations. The nonlinear camera and robot manipulator model is linearized and decoupled online by an adaptive algorithm. The adaptive controller is then computed based on the linearized model evaluated at current linearized point. The proposed solution is robust and easy to implement in different industrial robotic systems. Various scenarios are used in simulations to validate the effectiveness and robust performance of the proposed controller.

## 1. Introduction

Automatic alignment plays a crucial role in industrial assignments, such as micromanipulation, autonomous welding, and industrial assembly [1-3]. Visual servoing is a powerful tool that is commonly used in this application and guides the robotic systems to their desired poses [4-6]. The task to be solved by a visual servoing system is to provide velocities of the end effector that stabilizes and minimizes the difference between image features extracted by a vision device and the desired configurations [7].

The most known visual servoing techniques can be divided into two main categories: Image-Based Visual Servoing (IBVS) and Position-Based Visual Servoing (PBVS) [7]. IBVS is designed to drive the robot manipulators via feedback loops that matches 2D image features, such as points [7], lines [8], and entire images [9] while PBVS architecture extracts features in an image and estimates the pose, (including position and orientation) with respect to a 3D coordinate of frame in workspace and the difference between current pose and desired pose is defined as the control error [10]. In either case, an interaction matrix that relates the derivative of an image or pose features to the spatial velocities of the end effector must be computed [7][11].

In this research paper, the IBVS structure is explored and applied to a fastening and unfastening manufacturing scenario. Compared to PBVS, IBVS has the advantage of being more robust against camera parametric variations and noise but is more vulnerable to local minima and image singularities [7]. In general, the interaction matrix can be computed by using direct depth information [12-13] or by approximation of the depth [14-15] or by depth-independent interaction matrix [16]. However, these methods use redundant features to invert a non-square interaction matrix, as required in controller design, this in turn causes problems of local minima and image singularities. To address these issues, [17] uses the stereo visual system and provides a square interaction matrix. We have seen more trends in using stereo visual configuration in recent visual servoing works [18-19].

In most related works, a simple kinematic model of the camera is solely used in generating an interaction matrix [7, 20-21]. Dynamic visual servoing studied by [22-23, 17] includes the dynamic of robot arms in their models and they argue to have higher performance and enhanced stability in their response. However, controllers built in these papers are usually achieved with the PID control technique or its simplified variations. One improvement in this work is to use Youla robust control design technique [24] that includes both kinematics and dynamics in the model development stage. This advanced control technique can increase the robustness and stability of the system for high-speed tasks.

The eye-in-hand (EIH) vision system and the eye-to-hand (ETH) vision system are two kinds of camera configurations that have been widely used in visual servoing [18-19]. EIH has the freedom to obtain adequate environmental information, but the camera attached to the robot arms occupies more space and reduces flexibility of robot movement. On the other hand, in ETH systems, robot movements are not affected by the image extraction process. However, it usually suffers information loss and control loss of the robot when the point features are outside the view of the camera. One possible way to tackle this problem is to install several cameras to cover the whole workspace [19][25]. In this paper, we address this problem by introducing an innovative method of designing a feedforward-feedback control architecture. A designed feature estimation loop ensures stable and fast motion control when point features are outside the field of view. As 3D point features move inside the field of view, IBVS feedback loop preserves the precision of the pose at the end of the control period.

Adaptive control methods have been widely used in recent research on visual servoing fields. Liu and Wang [26] presented a new simplified adaptive controller for visual servoing of robot manipulators, which is based on the Slotine-Li algorithm [27]. They developed an adaptive algorithm to estimate unknown geometric parameters, such as the depths of the image features in the interaction matrix. In addition, in another paper [28], researchers explored a new resilient adaptive dynamic tracking control scheme for a fully uncalibrated IBVS system with unknown actuator faults. An effective adaptive algorithm was developed to estimate uncalibrated parameters in the camera, robot, and end-effector, which appear with a highly coupled and nonlinear form in the composite Jacobian matrix. Moreover, we have seen applications of adaptive controls with neural networks in visual servoings. For example, Qiu and Wu [29] have developed an adaptive neural network based IBVS dynamic method for both eye-in-hand and eye-to-hand camera configurations with unknown dynamics and external disturbances. In most of applications, adaptive algorithms are used to estimate unknown parameters in models. However, adaptive controller can also be used in dealing with coupled and nonlinear dynamic models in the IBVS structures, and this application is less addressed in literature. In this paper, one innovative contribution is to develop an adaptive feedback loop controller based on linear Youla parametrization to stabilize the nonlinear IBVS system in the whole range of operations. Simulation results for the various scenarios are presented and the robustness to noise and model uncertainties in the manufacturing process of fastening and unfastening are examined.

## 2. System Configuration

### 2.1. System Topology for fastening and unfastening scenarios

In this paper, an automatic alignment system is used to align a screwdriver, which is attached to the gripper of the robot arm and guide it to move above a screw at a prescribed location. The system is composed of a 6-DoF robot manipulator, a stereo camera, and the tools as shown in Figure 1. In this Figure, a camera is placed on the front of the workspace. The base frame {O} and the end-effector frame {E} are attached to the robot. A cartesian frame {C} is attached to the optical center of the camera. The body of the screwdriver is modeled as a cylinder and its central axis is approximately parallel to the Z-axis of the end-effector frame {E}.

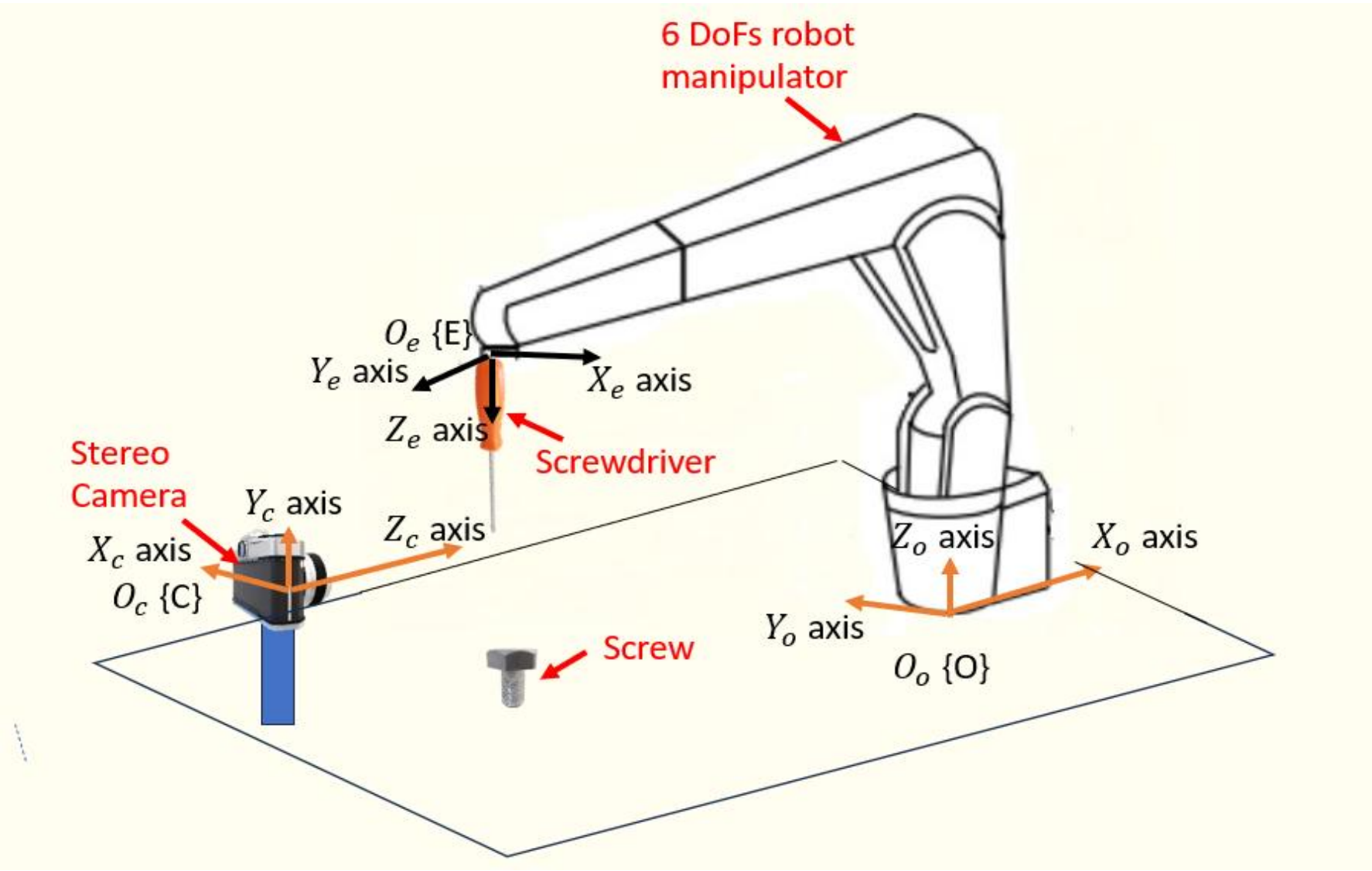

Figure 1. Configuration of the alignment system.

### 2.2. Point Feature Extraction

To capture the pose (position and orientation) of the screwdriver, two circular fiducial markers are placed on it. The Hough transformation [30] in computer vision can isolate circular markers in the image and localize the center of them. A general Hough transformation process is discussed below.

A circle with a radius $R$ and center $(a, b)$ can be defined in the following equation:

$$(x - a)^2 + (y - b)^2 = R^2 \tag{1}$$

Where $(x, y)$ are points on the perimeter of the circle. The three-element tuple $(a, b, R)$ uniquely parameterizes a circle in an image.

To provide an example for this mapping, we only consider one of the circles on our tool and explain the process as follows: the locus of $(a, b)$ points in the parameters space falls on a circle of radius $R$ centered at $(x, y)$, which are mapped to the parameters space and only three points are shown for this example. Each point in a geometric space generates a circle in the parameters space as shown in Figure 2. In case the geometric space points belong to the same circle as shown, the center of this circle will be represented by the intersection of circles in the parameters Space. During the calculation of the Hough transformation, each image point with the coordinates $(x, y)$ generates tuples $(a, b, R)$ with the unknown R. By creating different circles in geometric space, we can then generate different tuples. The tuple with the most intersections of the circles, in parameters space, provides the center and the radius of the center in the geometric space.

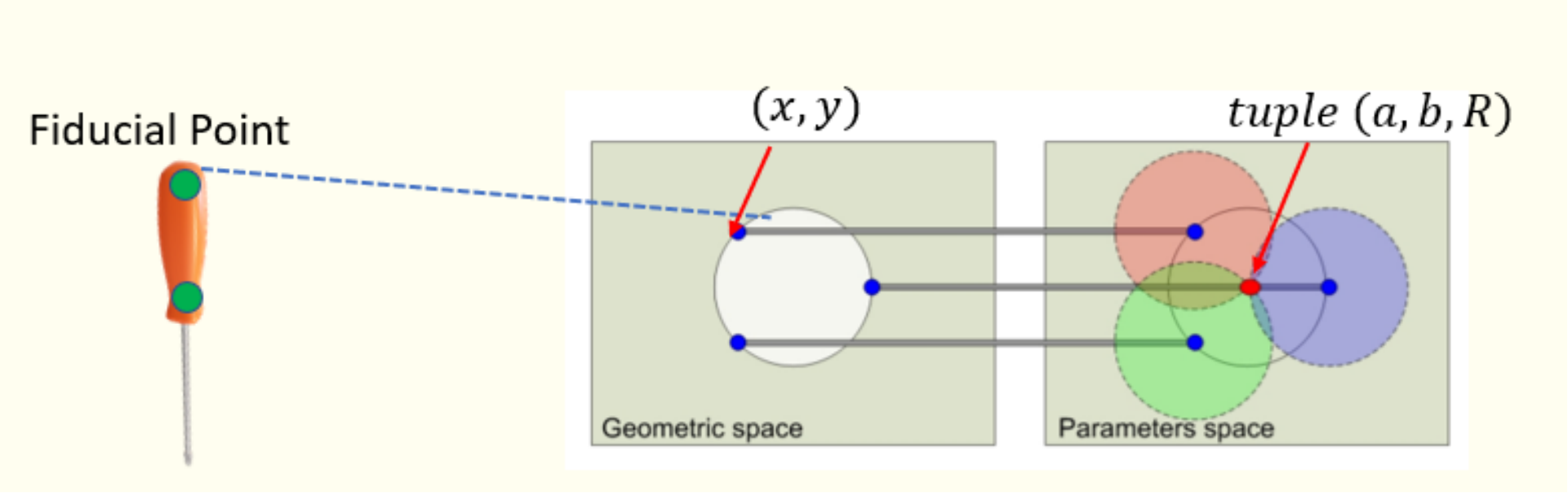

Figure 2. Feature extraction by Hough Transformation. (Note: the red point in parameters space represents a possible tuple $(a, b, R)$.

Then the stereo camera provides the image coordinates of the center of the two fiducial markers as:

$$\widehat{p_{I_1}^T} = [\widehat{ul_{I_1}}\ \widehat{ur_{I_1}}\ \widehat{v_{I_1}}\ ]^T \tag{2}$$

$$\widehat{p_{I_2}^T} = [\widehat{ul_{I_2}}\ \widehat{ur_{I_2}}\ \widehat{v_{I_2}}\ ]^T \tag{3}$$

Where $\widehat{p_{I_1}^T}$ and $\widehat{p_{I_2}^T}$ are image coordinates of circular centers of the first and the second fiducial markers. $\widehat{ul}$ and $\widehat{ur}$ are $u$ axis coordinates measured on the left and right lens's plane of the stereo camera, and $\hat{v}$ is the $v$ axis coordinates.

### 2.3. Control System Block Diagram

The feedback-feedforward pose alignment control block diagram with feature estimation feedback is given in Figure 3. The point features of fiducial markers extracted from images captured by the stereo camera. The current point features are denoted as $\widehat{p_I} = [\widehat{p_{I_1}^T}, \widehat{p_{I_2}^T}]$, and the target point features are denoted as $\overline{p_I} = [\overline{p_{I_1}^T}, \overline{p_{I_2}^T}]$. During the alignment, the coordinates of the target point features are fixed in the image frames, while the coordinates of the current point features vary with the movement of the end-effector.

The inner-loop joint control is an inner feedback loop (which is not shown in this Figure, and the detail of it will be discussed in the later section) that moves the robot arms to the target rotational angle $q_{Tref}$, which is commanded by the outer feedback controller and feedforward controller. Dynamic model of the robot manipulator is included in the inner-loop controlled system. Sensor noise may originate in the inner joint control loop from the low fidelity cheap encoder joint sensors and the dynamic errors from the joint, e.g., compliances. All sources of noise from the joint control loop are combined and modeled as an input disturbance, $d_{q_T}$, to the outer control loop.

The feedforward control loop is an open loop which brings the tool as close to the target position as possible in the presence of the input disturbance, $d_{q_T}$. The outputs of the feedforward are the reference joint angles of rotations, $q_{T_feedforward}$, which are added to the outer feedback controller outputs, $q_{T_feedback}$ , and set as the targets for the joint control inner-loop. The function of the Tool-on-Robot kinematics model is to transform a set of current joint angles of the tool manipulator to the current pose of the tool on the end-effector using a kinematic model of the robot arm.

The movement of the tool can be adjusted by the adaptive feedback control loop. The adaptive controller is computed online according to the current image coordinate $\hat{p}_I$. The feedback control loop rejects the input disturbance, $d_{q_T}$, by minimizing the error between the target coordinates of two makers, $\bar{p}_I$, and the image estimation, $\hat{p}_I$ in the image frame.

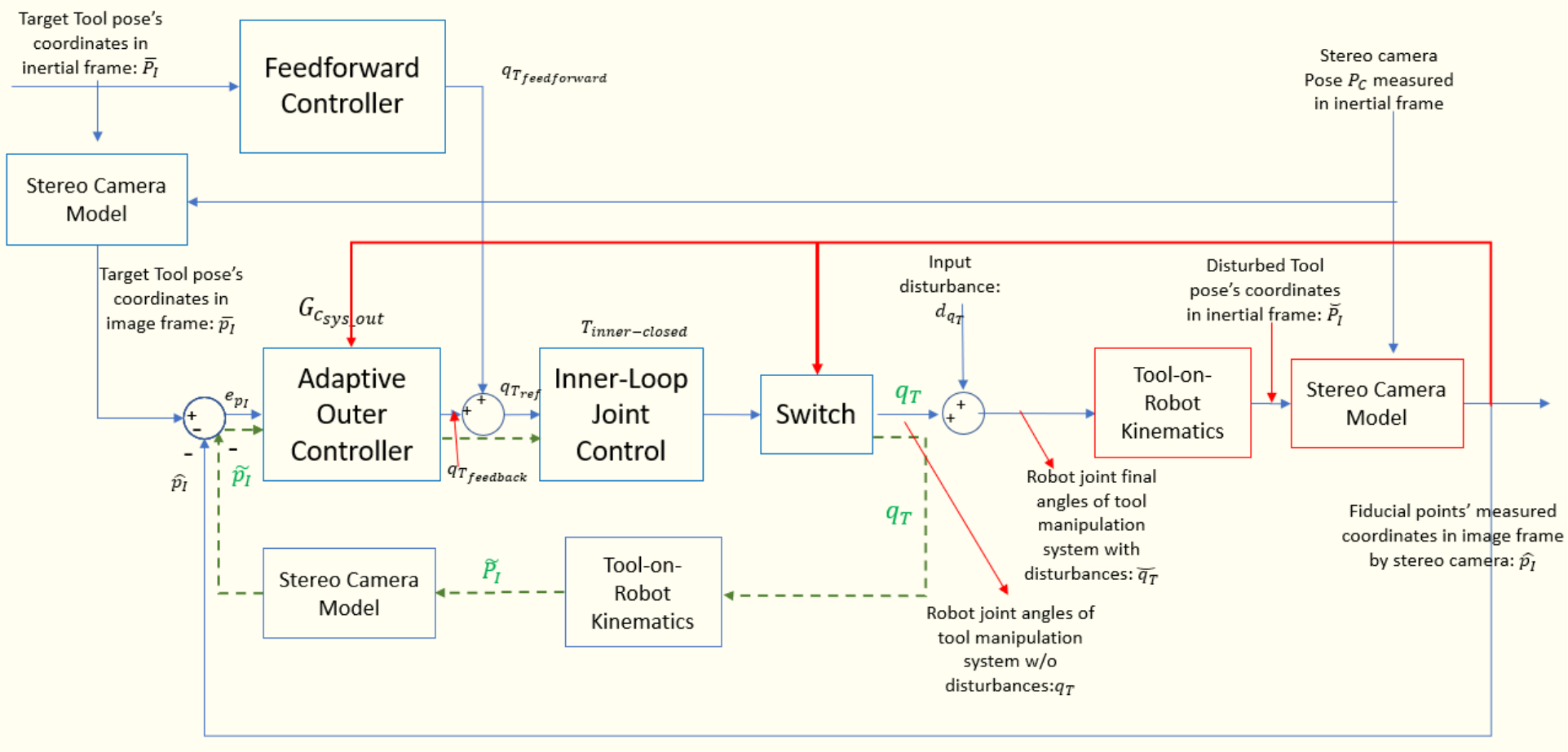

Figure 3. Block diagram of Feedforward-Feedback control loop with feature estimation (Note: Blocks in red lines are referred to HIL Models)

The feedforward and feedback controllers work simultaneously to move the tool to the target pose location in the tool manipulation system. The combined target $q_{T_{ref}}$ are the inputs to the joint control loop so that both controllers manipulate the tool pose. The benefit of designing both feedback and feedforward controls for the manipulation system is to reduce the time duration. If only feedback control is utilized, the pose estimation generated from the visual system requires image processing and makes the tool movement very slow. We can divide the task of the tool movement control into two stages. In the first stage, under the action of the feedforward control, the tool moves to an approximate location that is close to the desired destination. In the second stage, the feedback controller moves the tool to the precise target location using the tool pose estimation from the camera. In addition, the camera has a range of view and can only detect the tool and measure its 2D feature as $\hat{p}_I$ when it is not far away from the target. When the tool is moving from a location that is not in the camera range of view, we must estimate the feature as $\tilde{p}_I$ until the tool moves into the range of view.

We developed another feedback loop so that we can estimate the 2D coordinates of the tool from the same mathematical model similar to the Hardware-In-Loop (HIL) models when the real measurements are unavailable. In Figure 3, the normal feedback loop (blue lines) is preserved when the tool is inside the camera range of view and hence, the camera can measure the tool's 2D feature $\hat{p}_I$. However, when the tool is outside the range of view, the 2D feature can only be estimated as $\tilde{p}_I$ (green dashed lines). We can implement a bump-less switch to smoothly switch between these modes of operations. The switching signal is activated when the tool moves in or out of the camera range of view (or when the camera detects fiducial point features).

## 3. Model Development

### 3.1 Tool-on-Robot Kinematics

The robotic model we use in this article is a specific manipulator ABB IRB 4600 [31], which is an elbow manipulator with spherical wrist as shown in Figure 4. The physical dimension parameters of this robot manipulator $(a_1, L_1, L_2, L_3, L_4, L_t)$ are summarized in the Appendix section. This model of robot has a total of 6 links with three composed of the robot arms and other three composed of wrists. A joint is connected between each of the two adjacent links and there is a total of 5 convolutional joints. In addition to the base of the robot arm, we have attached to each joint a cartesian coordinate, as shown in Figure 4. Joint axes $Z_0$, ⋯ $Z_5$ are the rotational direction of each joint. Their rotational angles are defined as $\theta_1$, ⋯ $\theta_6$. The benefit of spherical wrist is that the wrist joint center (where $Z_3$, $Z_4$ and $Z_5$ intersect) is kept stationary whichever the wrist orients. Therefore, the task of orientation will not affect the position of the wrist center. Each coordinate attached on the frame is generated based on the procedures that derive the forward kinematics by Denavit-Hartenberg convention (or D-H convention) [32]. In the coordinate frame $O_6X_6Y_6Z_6$ attached to the end-effector, set the unit vector $\widehat{k_6} = \hat{a}$ along the direction of $z_6$. Set $\widehat{j_6} = \hat{s}$ in the direction of gripper closure and set $\widehat{\imath_6} = \hat{n}$ as $\hat{s} \times \hat{a}$. The values of $\hat{a}, \hat{s}$ *and* $\hat{n}$ in the base frame $O_0X_0Y_0Z_0$ define the orientation of the object (tool) mounted at the end-effector.

In D-H convention, each homogeneous transformation matrix $\boldsymbol{A_i^{i-1}}$ (from frame $\boldsymbol{i}$ to frame $\boldsymbol{i-1}$) can be represented as a product of four basic transformations:

$$\boldsymbol{A_i^{i-1}} = Rot_{z,\theta_i} Trans_{z,d_i} Trans_{x,a_i} Rot_{x,\alpha_i} \quad (4)$$

$$= \begin{bmatrix} c_{\theta_i} & -s_{\theta_i} & 0 & 0 \\ s_{\theta_i} & c_{\theta_i} & 0 & 0 \\ 0 & 0 & 1 & 0 \\ 0 & 0 & 0 & 1 \end{bmatrix} * \begin{bmatrix} 1 & 0 & 0 & 0 \\ 0 & 1 & 0 & 0 \\ 0 & 0 & 1 & d_i \\ 0 & 0 & 0 & 1 \end{bmatrix} * \begin{bmatrix} 1 & 0 & 0 & a_i \\ 0 & 1 & 0 & 0 \\ 0 & 0 & 1 & 0 \\ 0 & 0 & 0 & 1 \end{bmatrix} * \begin{bmatrix} 1 & 0 & 0 & 0 \\ 0 & c_{\alpha_i} & -s_{\alpha_i} & 0 \\ 0 & s_{\alpha_i} & c_{\alpha_i} & 0 \\ 0 & 0 & 0 & 1 \end{bmatrix}$$

$$= \begin{bmatrix} c_{\theta_i} & -s_{\theta_i} c_{\alpha_i} & s_{\theta_i} s_{\alpha_i} & a_i c_{\theta_i} \\ s_{\theta_i} & c_{\theta_i} c_{\alpha_i} & -c_{\theta_i} s_{\alpha_i} & a_i s_{\theta_i} \\ 0 & s_{\alpha_i} & c_{\alpha_i} & d_i \\ 0 & 0 & 0 & 1 \end{bmatrix} \quad (5)$$

Note: $c_{\theta_i} \equiv \cos(\theta_i), c_{\alpha_i} \equiv \cos(\alpha_i), s_{\theta_i} \equiv \sin(\theta_i), s_{\alpha_i} \equiv \sin(\alpha_i)$ (6)

Where $\theta_i$, $a_i$, $\alpha_i$ and $d_i$ are parameters of link $i$ and joint $i$, $a_i$ is the link length, $\theta$ is the rotational angle, $\alpha_i$ is the twist angle and $d_i$ is the offset length between the previous $(i-1)^{th}$ and the current $i^{th}$ robot links. The quantities of each parameter $q_i$, $a_i$, $\alpha_i$ and $d_i$ in (5) are calculated in Table 1 based on the steps in [32].

Table 1. DH-Parameter for elbow manipulator with spherical wrist

| Link | $a_i$ | $\alpha_i$(rad) | $d_i$ | $\theta_i$(rad) |
|---|---|---|---|---|
| 1 | $a_1$ | $-\pi/2$ | $L_1$ | $\theta_1^*$ |
| 2 | $L_2$ | 0 | 0 | $\theta_2^* - \pi/2$ |
| 3 | $L_3$ | $-\pi/2$ | 0 | $\theta_3^*$ |
| 4 | 0 | $\pi/2$ | $L_4$ | $\theta_4^*$ |
| 5 | 0 | $-\pi/2$ | 0 | $\theta_5^*$ |
| 6 | 0 | 0 | $L_t$ | $\theta_6^* + \pi$ |

Only the angle are variables and shown with *

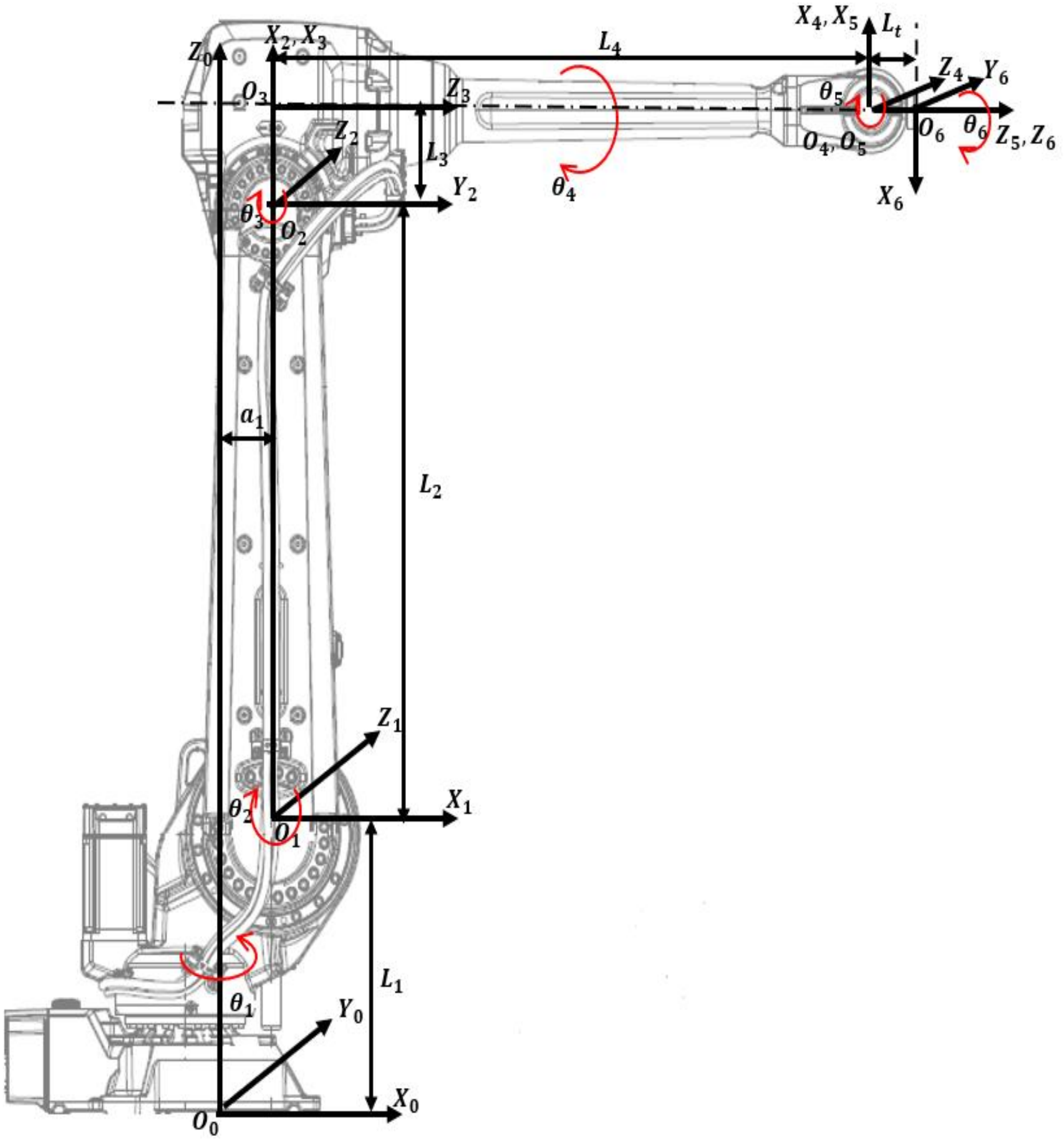

Figure 4. IRB ABB 4600 Model with attached frames

We can generate the transformation matrix $\boldsymbol{T_E^O}$ from the base inertial frame $\boldsymbol{O_0X_0Y_0Z_0}$ ($\boldsymbol{P^o}$) to the end-effector frame $\boldsymbol{O_6X_6Y_6Z_6}$ ($\boldsymbol{P^E}$) :

$$T_E^O = A_1^0 A_2^1 A_3^2 A_4^3 A_5^4 A_6^5 \tag{7}$$

Assume a screwdriver with a length of $L_{tool}$ is grasped by the gripper at the end effector, and the body of screwdriver is parallel to the $Z_6$ axis. Two fiducial feature points attached to it are placed on the screwdriver where their coordinates measured in the end-effector frame as $\widetilde{P_{I_1}}\{E\} = [0,0,0]^T$, and $\widetilde{P_{I_2}}\{E\} = [0,0,L_{tool}/2]^T$. Then those coordinates of points in the frame $\boldsymbol{P}^E$ can be transformed to the frame $\boldsymbol{P}^O$ as:

$$\widetilde{P_{I_1}}\{O\} = T_E^O\ \widetilde{P_{I_1}}\{E\} \tag{8}$$

$$\widetilde{P_{I_2}}\{O\} = T_E^O\ \widetilde{P_{I_2}}\{E\} \tag{9}$$

To be consistent with notations in the control block diagram (Figure 3), the disturbed angles of rotation in transformation matrix $T_E^O$ should be written as $\widetilde{q_T} = [\theta_1^*, \theta_2^*, \theta_3^*, \theta_4^*, \theta_5^*, \theta_6^*]^T$.

Take $\breve{P}_I\{E\} = [\breve{P}_{I_1}\{E\}, \breve{P}_{I_2}\{E\}]$, $\breve{P}_I\{O\} = [\breve{P}_{I_1}\{O\}, \breve{P}_{I_2}\{O\}]$, the kinematics model can be written as a nonlinear function as follows,

$$\breve{P}_I\{O\} = T_E^O(\breve{q}_T) \cdot \breve{P}_I\{E\} \quad (10)$$

### 3.2 Dynamic models of 6 DoFs Robot Manipulator

Dynamic models are included in the inner control loop in Figure 3. The dynamic model of a serial of 6-link rigid, non-redundant, fully actuated robot manipulator can be written as:

$$(D(q) + J)\ddot{q} + (C(q, \dot{q}) + \frac{B}{r})\dot{q} + g(q) = u \quad (11)$$

Where $q \epsilon \mathbb{R}^{6X1}$ is the vector of joint positions, and $u \epsilon \mathbb{R}^{6X1}$ is the vector of electrical power input from DC motors inside joints, $D(q) \epsilon \mathbb{R}^{6X6}$ is the symmetric positive defined matrix, $C(\boldsymbol{q}, \dot{\boldsymbol{q}}) \epsilon \mathbb{R}^{6X6}$ is the vector of centripetal and Coriolis effects, $g(\boldsymbol{q}) \epsilon \mathbb{R}^{6X1}$ is the vector of gravitational torques, $J \epsilon \mathbb{R}^{6X6}$ is a diagonal matrix expressing the sum of actuator and gear inertias, $B \epsilon \mathbb{R}^{6X1}$ is the damping factor, $r \epsilon \mathbb{R}^{6X1}$ is the gear ratio.

### 3.3 Stereo Camera Model

Assume the camera's position and orientation is fixed and pre-known in the inertial frame. The transformation matrix $T_C^O$ from the camera attached cartesian frame {C} to the inertial base frame {O} is:

$$T_C^O = \begin{bmatrix} n_x & s_x & a_x & d_x \\ n_y & s_y & a_y & d_y \\ n_z & s_z & a_z & d_z \\ 0 & 0 & 0 & 1 \end{bmatrix} \quad (12)$$

Where $[n_x, n_y, n_z]^T$, $[s_x, s_y, s_z]^T$ and $[a_x, a_y, a_z]^T$ are the camera's directional vector of Yaw, Pitch and Roll in the base frame $O_0X_0Y_0Z_0$ . And $[d_x, d_y, d_z]^T$ are the vector of absolute position of the center of the camera in the base frame $O_0X_0Y_0Z_0$.

The transformation matrix $T_O^C$ from the inertial base frame {O} to the camera frame {C} can be derived as:

$$T_O^C = T_C^{O^{-1}} \quad (13)$$

Then the coordinates of fiducial points in frame $\{O\}$ as expressed in Eq. (8) and Eq. (9) can be transformed to the frame {C} as:

$$\breve{P}_{I_1}\{C\} = T_O^C\, \breve{P}_{I_1}\{O\} \quad (14)$$

$$\breve{P}_{I_2}\{C\} = T_O^C\, \breve{P}_{I_2}\{O\} \quad (15)$$

In order to detect depth of an object, a stereo camera is required for implementation. As shown in Figure 5, two identical cameras are separated by a baseline distance b. In this paper, the pinhole camera model is used to represent each camera. A 3D point $\breve{P}_I\{C\} = [X_I^C, Y_I^C, Z_I^C]^{\boldsymbol{T}}$, measured in the camera cartesian frame {C}, is projected to two parallel virtual image planes, and each plane is located between each optical center ($C_l$ or $C_R$) and the object point $\breve{P}_I$. Denote the coordinates of the point projected on the left image plane as $[ul\ v]^T$ and the coordinates of the point projected on the right image plane as $[ur\ v]^T$.

The relationship between the image coordinates $(ul\ v)$, $(ur\ v)$, and the coordinates $(X_I^C, Y_I^C, Z_I^C)$ on the normalized imaging plane is given by:

$$\begin{bmatrix} \boldsymbol{u_l} \\ \boldsymbol{v_l} \\ \boldsymbol{1} \end{bmatrix} = \frac{\boldsymbol{1}}{\boldsymbol{Z^C}} \begin{bmatrix} \boldsymbol{f_u} & \boldsymbol{s_c} & \boldsymbol{u_0} \\ \boldsymbol{0} & \boldsymbol{f_v} & \boldsymbol{v_0} \\ \boldsymbol{0} & \boldsymbol{0} & \boldsymbol{1} \end{bmatrix} \begin{bmatrix} \boldsymbol{X_I^C} \\ \boldsymbol{Y_I^C} \\ \boldsymbol{Z_I^C} \end{bmatrix} - \frac{\boldsymbol{b}}{\boldsymbol{2Z^C}} \begin{bmatrix} \boldsymbol{f_u} \\ \boldsymbol{0} \\ \boldsymbol{0} \end{bmatrix} \quad (16)$$

$$\begin{bmatrix} u_r \\ v_r \\ 1 \end{bmatrix} = \frac{1}{Z^C}\begin{bmatrix} f_u & s_c & u_0 \\ 0 & f_v & v_0 \\ 0 & 0 & 1 \end{bmatrix}\begin{bmatrix} X_I^C \\ Y_I^C \\ Z_I^C \end{bmatrix} + \frac{b}{2Z^C}\begin{bmatrix} f_u \\ 0 \\ 0 \end{bmatrix} \tag{17}$$

Where $\boldsymbol{f_u}$ and $\boldsymbol{f_v}$ are the horizontal and the vertical focal lengths, and $\boldsymbol{s_c}$ is a skew coefficient. In most cases, $\boldsymbol{f_u}$ and $\boldsymbol{f_v}$ are different if the image horizontal and vertical axes are not perpendicular. In order not to have negative pixel coordinates, the origin of the image plane will be usually chosen at the upper left corner instead of the center. $\boldsymbol{u_0}$ and $\boldsymbol{v_0}$ describe the coordinate offsets.

When there are no coordinate offsets and skews between u-v image plane, that is $\boldsymbol{s_c} = \boldsymbol{u_0} = \boldsymbol{v_0} = \boldsymbol{0}$, the above relationships between 3D domain to 2D image domains of a point $\boldsymbol{P}$ in the workspace can be simplified as:

From 3D to 2D:

$$v = v_l = v_r = \frac{\boldsymbol{Y_I^C}}{Z_I^C} f_v \tag{18}$$

$$u_l = \frac{2\boldsymbol{X_I^C} - b}{2Z_I^C} f_u \tag{19}$$

$$u_r = \frac{2\boldsymbol{X_I^C} + b}{2Z_I^C} f_u \tag{20}$$

Note $v_l$ and $v_r$ have the same value and they are denoted as the same parameter $v$.

Finally, from stereo camera model equations above, the coordinates of two fiducial points in the image frame can be expressed as:

$$\widehat{\boldsymbol{p_{I_1}}} = \left[\frac{\boldsymbol{2X_{I_1}^C - b}}{\boldsymbol{2Z_{I_1}^C}}\boldsymbol{f_u}, \frac{\boldsymbol{2X_{I_1}^C + b}}{\boldsymbol{2Z_{I_1}^C}}\boldsymbol{f_u}, \frac{\boldsymbol{Y_{I_1}^C}}{\boldsymbol{Z_{I_1}^C}}\boldsymbol{f_v}\right]^{\boldsymbol{T}} = [\boldsymbol{ul_{I_1}}\ \boldsymbol{ur_{I_1}}\ \boldsymbol{v_{I_1}}]^{\boldsymbol{T}} \tag{21}$$

$$\widehat{p_{I_2}} = \left[\frac{2X_{I_2}^C - b}{2Z_{I_2}^C}f_u, \frac{2X_{I_2}^C + b}{2Z_{I_2}^C}f_u, \frac{Y_{I_2}^C}{Z_{I_2}^C}f_v\right]^T = [ul_{I_2}\ ur_{I_2}\ v_{I_2}]^T \tag{22}$$

Where $\breve{P}_{I_1}\{C\} = [X_{I_1}^C, Y_{I_1}^C, Z_{I_1}^C]$, and $\breve{P}_{I_2}\{C\} = [X_{I_2}^C, Y_{I_2}^C, Z_{I_2}^C]$ are coordinates of two fiducial points measured in the camera's frame {C}.

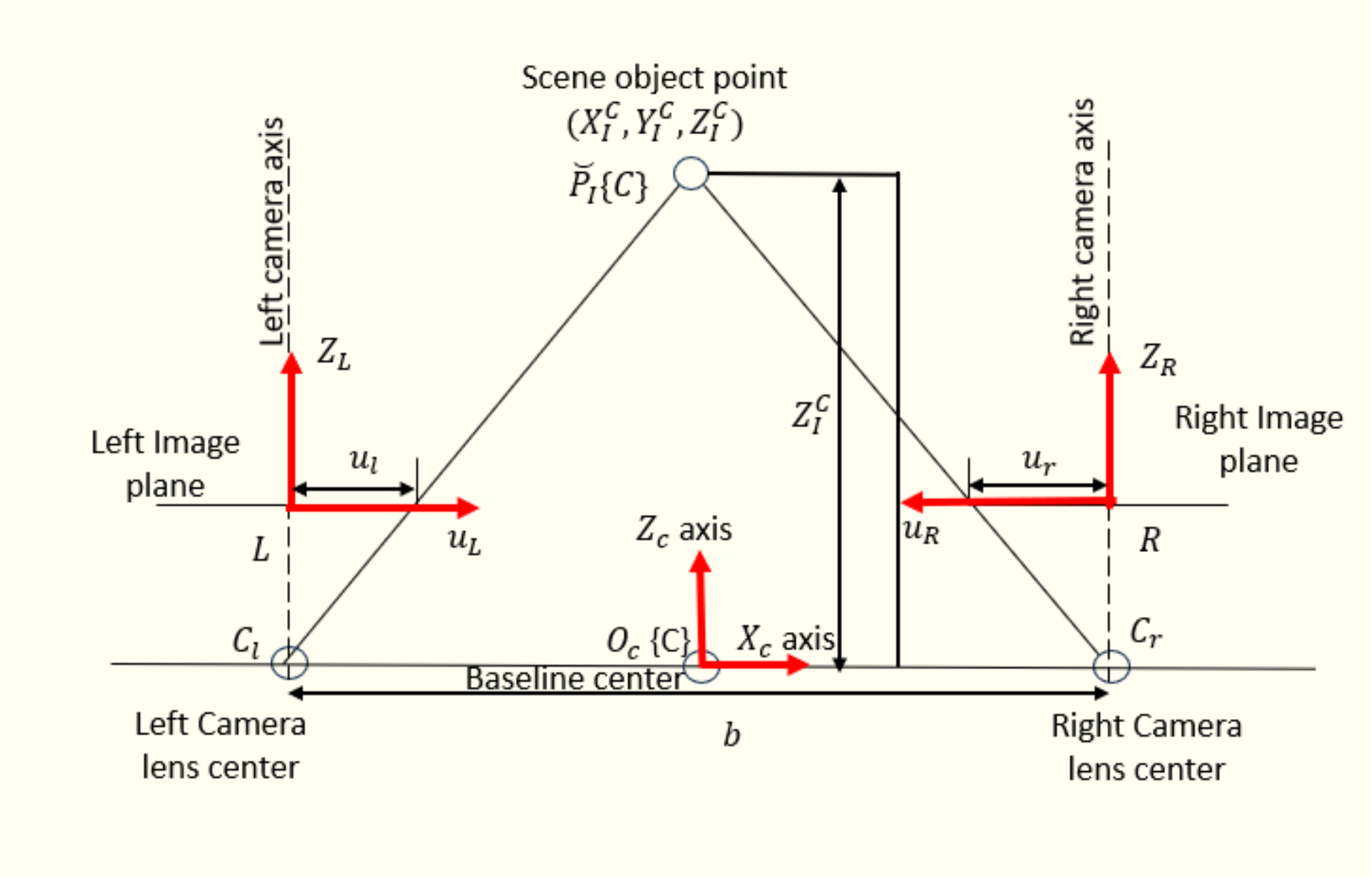

Figure 5. The projection of a scene object on the stereo camera's image planes. Note: $v$ coordinate on each image plane is not shown in the plot but is measured along the axis that is perpendicular to and point out of the plot.

A specific stereo camera model: Zed 2 [33] is used in the model development and simulations. The parameters of this type are summarized in the Appendix.

## 4. Control Strategy

### 4.1 Inner-loop joint control

Our previous paper [34] has provided details of developing the inner-loop joint controller with Youla feedback linearization. Here in this section, we only discuss a summary of the design.

Eq. (11) expresses the dynamic model of a 6 DoFs robot manipulator. We simplify this equation as follows:

$$M(q)\ddot{q} + h(q,\dot{q}) = \mathrm{u} \tag{23}$$

with

$$M(q) = D(q) + J \tag{24}$$

$$h(q,\dot{q}) = (C(q,\dot{q}) + \frac{B}{r})\dot{q} + \mathrm{g}(q) \tag{25}$$

Then, transform the control input as following:

$$\mathrm{u} = M(q)v + h(q,\dot{q}) \tag{26}$$

where $v$ is a virtual input. Then, substitute for u in Eq. (23) using Eq. (26), and since $M(q)$ is invertible, we will have a reduced system equation as follows:

$$\ddot{q} = v \tag{27}$$

By using feedback linearization with Youla parameterization, we can design the controller system as shown in Figure 6.

The Joint controller is designed as:

$$G_{c_{sys_inner}} = \frac{3{\tau_{in}}^2 s + 1}{{\tau_{in}}^3 s + 3{\tau_{in}}^2} \cdot I_{6\times 6} \tag{28}$$

And the closed loop transfer function can be expressed as:

$$T_{sys_inner} = \frac{(3\tau_{in}s + 1)}{(\tau_{in}s + 1)^3} \cdot I_{6\times 6} \tag{29}$$

Where $I_{6\times 6}$ is a 6x6 identity matrix, and $\tau_{in}$ specifies the pole and zero locations and represents the bandwidth of the control system. $\tau_{in}$ can be tuned so that the response can be fast with less-overshoot.

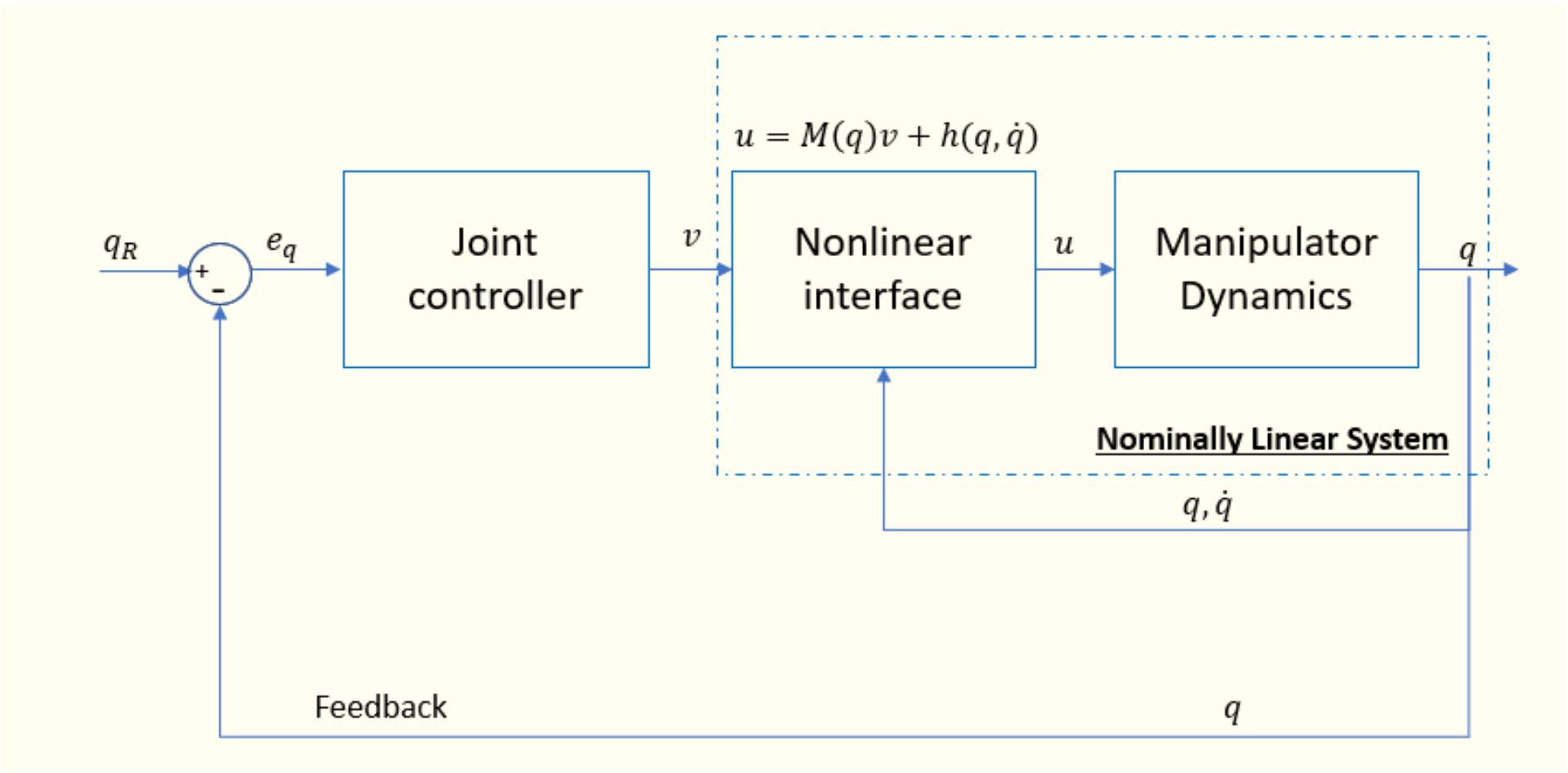

Figure 6. Block diagram of inner joint control loop with feedback linearization.

### 4.2 Adaptive Controller Design

In control system block diagram (Figure 3), the linear closed loop transfer function of inner-loop joint control is derived in Eq. (29). The tool-on-robot kinematics model and stereo camera model, which can be combined as a new model: camera-and-robot model, compose of the nonlinear part of the plant in outer feedback loop. Combining the equations in section 3, we can derive a nonlinear function denoted as $\mathcal{F}$ that relates the disturbed robot joint angles $\widetilde{q_T}$ and 2D image coordinates of two fiducial points $\widehat{p_I}$:

$$\widehat{p_I} = \mathcal{F}(\widetilde{q_T}) \tag{30}$$

We can use model linearization method by linearizing the nonlinear function (30) at different linearized points and design linear controllers utilizing these linearized models. By chosen a linearized point $\widetilde{q_T}^0$, the nonlinear function (30) can be linearized using the Jacobian matrix form as:

$$\widehat{p_I} = J(\widetilde{q_T}^0)\widetilde{q_T} + \mathcal{F}(\widetilde{q_T}^0) \tag{31}$$

Where $J(\widetilde{q_T}^0) \epsilon\ \mathbb{R}^{6X6}$ is the Jacobian matrix of $\mathcal{F}(\widetilde{q_T})$ evaluated as $\widetilde{q_T} = \widetilde{q_T}^0$.

Assuming $C_1 = J(\widetilde{q_T}^0)$, $C_2 = \mathcal{F}(\widetilde{q_T}^0)$, therefore, Eq. (31) can be rewritten as:

$$\widehat{p_I} = C_1\widetilde{q_T} + C_2 \tag{32}$$

Let's define $\widehat{p_I}' = \widehat{p_I} - C_2$ , then, the overall block diagram of the linearized system is shown in Figure 7.

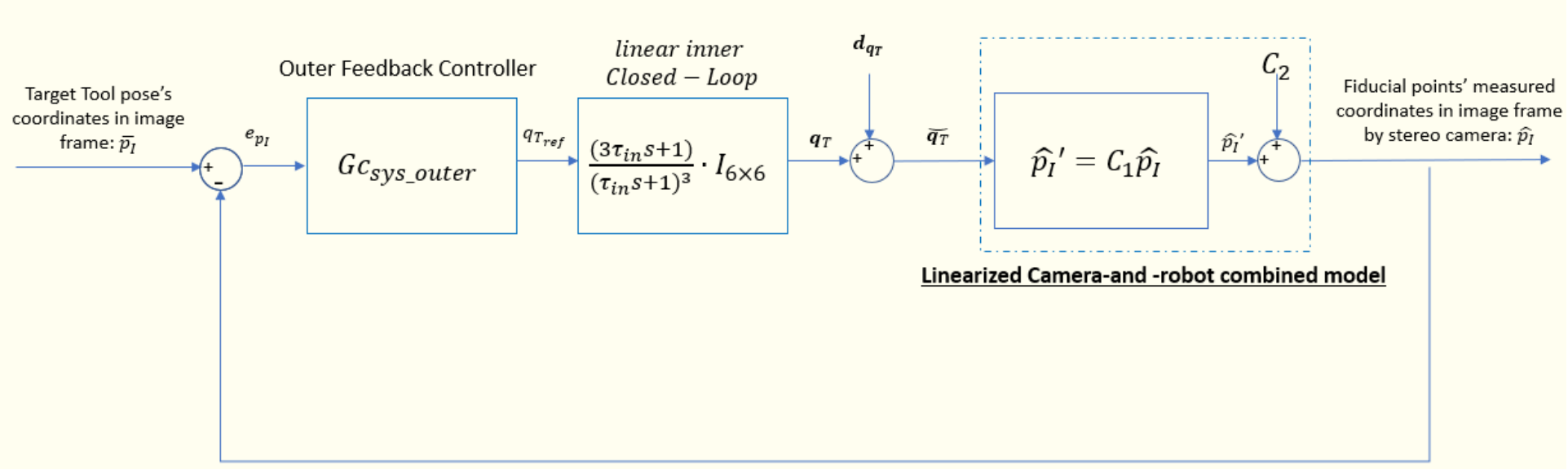

Figure 7. Block diagram of feedback loop with linearizated camera-and-robot combined model.

The linearized plant transfer function is derived as:

$$G_{p_{sys_outer}}^{\ linear} = \frac{\widehat{p_I}'}{q_{V_{ref_1}}} = C_1 \frac{(3\tau_{in}s + 1)}{(\tau_{in}s + 1)^3} \cdot I_{6\times6} \tag{33}$$

As $C_1$ is coupled, the first step to derive a controller for the multivariable system using model linearization is to find the Smith-McMillan form of the plant [35].

To obtain the Smith-McMillan form, we can decompose $G_p$ using singular value decomposition (SVD) as:

$$G_{p_{sys_outer}}^{\ linear} = U_L M_p U_R \tag{34}$$

where $U_L \epsilon\ \mathbb{R}^{6X6}$ and $U_R\ \epsilon\ \mathbb{R}^{6X6}$are the left and right unimodular matrices, and $M_p$ $\epsilon\ \mathbb{R}^{6X6}$ is the Smith-McMillan form of $G_{p_{sys_outer}}^{\ linear}$. The SVD in this case only works for the replacement of Smith-McMillan form computation because we are dealing with a constant matrix and a dialogized transfer function matrix.

$M_p$ is a diagonalized transfer function matrix with each nonzero entry equals to a gain multiple the transfer function $\frac{(3\tau_{in}s+1)}{(\tau_{in}s+1)^3}$ ; For the $i^{th}$ row of $M_p$ the entry on the diagonal is:

$$M_p(i,i) = gain(i) \cdot \frac{(3\tau_{in}s+1)}{(\tau_{in}s+1)^3}, i\epsilon(1,2,3,4,5,6) \tag{35}$$

Where $gain\ \epsilon\ \mathbb{R}^{6X1}$ is a numerical vector.

The design of a Youla controller for each nonzero entry in $M_p$ is trivial in this case as all poles/zeros of the plant transfer function matrix are in the left half-plane, and therefore, they are stable. In this case, the selected decoupled Youla transfer function matrix, $M_Y$ , can be selected to shape the decoupled closed loop transfer function matrix, $M_T$. All poles and zeros in the original plant can be cancelled out and new poles and zeros can be added to shape the closed-loop system. Let's select a Youla transfer function matrix so that the decoupled closed-loop SISO system behaves like a second order Butterworth filter, such that:

$$M_T = \frac{{\omega_n}^2}{(s^2 + 2\zeta\omega_n s + {\omega_n}^2)} \cdot I_{6\times6} \tag{36}$$

where $\omega_n$ is called natural frequency and approximately sets the bandwidth of the closed–loop system. It must be ensured that the bandwidth of the outer-loop is smaller than the inner-loop, i.e., $1/\omega_n > \tau_{in}$. Parameter, $\zeta$ , is called the damping ratio, which is another tuning parameter.

We can then compute the decoupled diagonalized Youla transfer function matrix $M_Y$. The diagonal entry of $i^{th}$ row is denoted as $M_Y(i,i)$:

$$M_Y(i,i) = \frac{M_T(i,i)}{M_p(i,i)} = \frac{1}{gain(i)} \frac{{\omega_n}^2}{(s^2+2\zeta\omega_n s+{\omega_n}^2)} \frac{(\tau_{in}s+1)^3}{(3\tau_{in}s+1)},, i\epsilon(1,2,3,4,5,6) \tag{37}$$

The final coupled Youla, closed loop, sensitivity, and controller transfer function matrices are computed as:

$$Y_{sys_outer}^{linear} = U_R M_Y U_L \tag{38}$$

$$T_{y_{sys_outer}}^{\ linear} = G_{p_{sys_outer}}^{\ linear} \cdot Y_{sys_outer}^{linear} \tag{39}$$

$$S_{y_{sys_outer}}^{\ linear} = 1 - T_{y_{sys_outer}}^{\ linear} \tag{40}$$

$$G_{C_{sys_outer}}^{\ linear} = Y_{sys_outer}^{linear} \cdot (S_{y_{sys_outer}}^{\ linear})^{-1} \tag{41}$$

The controller developed in the above section is based on the linearization of the combined model at a particular linearized point $\widetilde{q_T}^0$. This controller can only stabilize at certain range of joint angles around $\widetilde{q_T}^0$. As current joint angles $\widetilde{q_T}$ deviates from $\widetilde{q_T}^0$, the error between the estimated linearized system (33) and the true nonlinear system (30) increases.

To tackle this problem, we develop an adaptive controller that is computed online based on linearization of the model at current joint angles. This control process is depicted in Figure 8.

The first step is to estimate current joint angles $\widetilde{q_T}$ from current measured images coordinates $\widehat{p_I}$. The mathematic models of stereo camera and robot kinematics have been given in the above sections, and a combined model expression is defined in (30). Therefore, the mathematical function of the inverse model can be derived and expressed as:

$$\widetilde{q_T} = \mathcal{F}^{-1}(\widehat{p_I}) \tag{42}$$

Where $\mathcal{F}^{-1}$ is the inverse function of (30). Given estimated current angle $\widetilde{q_T}$, we can calculate the Jacobian matrix of the nonlinear model at current time. By obtaining left and right unimodular matrices and Smith-McMillan form from singular value decomposition, the current linear controller $G_{C_{sys_outer}}^{linear}$ can be built by steps (35)-(41).

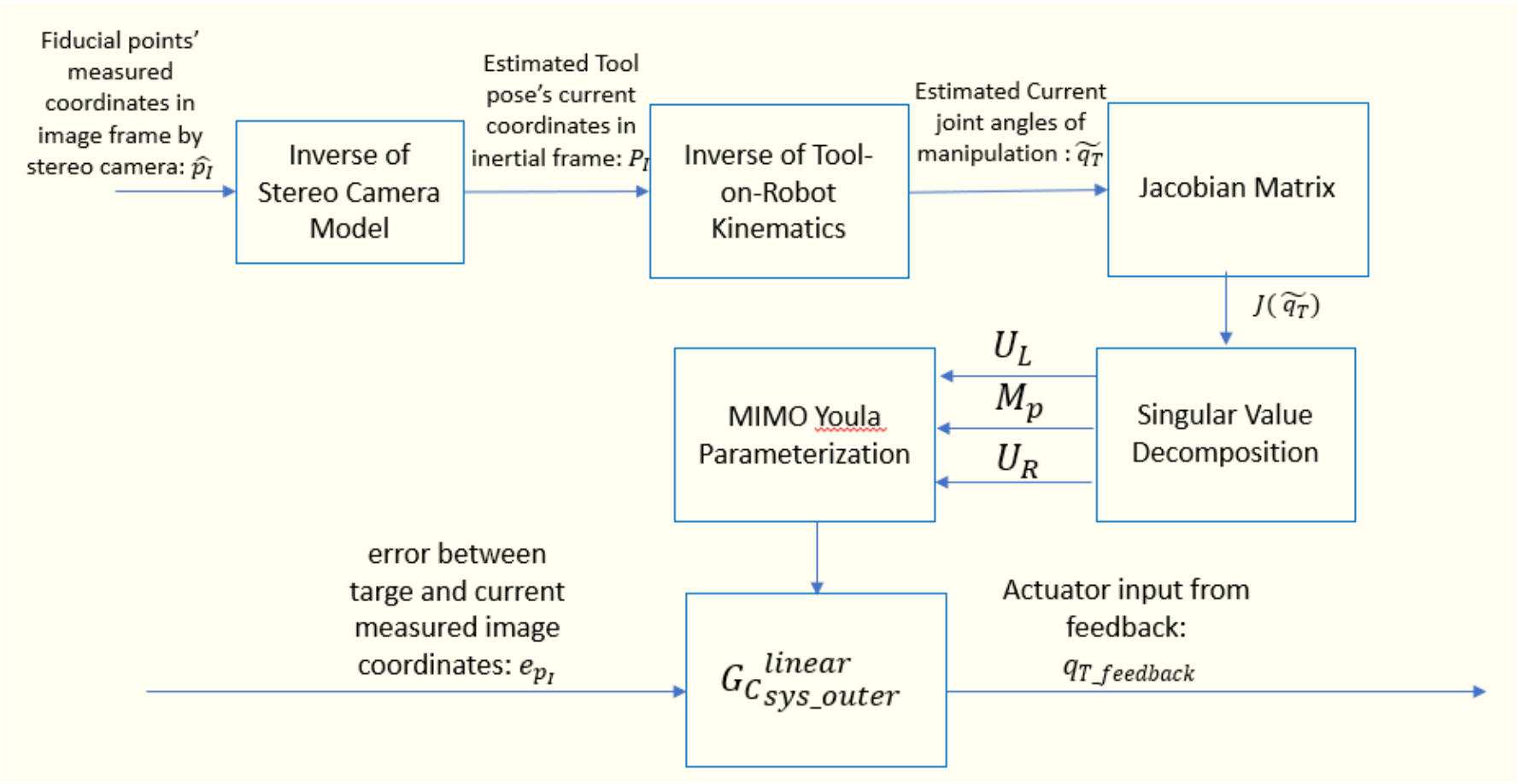

Figure 8. Block diagram of adaptive control design

In this way, the adaptive outer-loop controller is computed online based on current linearized plane model, which is evaluated at current estimated robot's joint angles. The mapping between nonlinear plant model and linearized model is complete in all joint angles' configurations, thus the adaptive feedback controller is robust in the entire range of operations.

### 4.3 Feedforward Control

Feedforward Controller generates a target rotational angle $q_{T_feedforward}$ based on the tool's coordinates at the target location. As shown in the Figure below, Feedforward controller contains two inverse processes of the original plant in the development. The first diagram illustrates the inverse process of the tool-on-robot kinematics model and the second diagram $T_{forward}$ shows the inverse process of inner-loop closed transfer function $T_{inner-loop}$.

Eq. (10) provides the mathematical expression of the tool-on-robot kinematics model. The inverse of (10) gives an expression of $\widetilde{q_T}$ from $\breve{P}_I\{O\}$ and $\breve{P}_I\{E\}$. We can define the inverse of tool-on-robot kinematics as follows:

$$\widetilde{q_T} = \mathcal{H}(\breve{P}_I\{O\}, \breve{P}_I\{E\}) \quad (43)$$

This process is so-called inverse kinematics of the robot manipulator, which finds the joint configurations given the coordinates of points in both the bottom frames and the end-effector frame.

The $T_{forward}$ can be designed as,

$$T_{forward} = \frac{1}{T_{inner-closed}} \frac{1}{(\tau_{forward}s+1)^2} = \frac{(\tau_{in}s+1)^3}{(3\tau_{in}s+1)} \frac{1}{(\tau_{forward}s+1)^2} \cdot I_{6\times 6} \quad (44)$$

The double poles s = -1/ $\tau_{forward}$ are added to make $T_{forward}$ proper. Choose $\tau_{forward}$ so that the added double poles are 10 times larger than the bandwidth of the original improper $T_{forward}$. In other words, $\tau_{forward}$ is chosen as,

$$\tau_{forward} = 0.1\tau_{in} \quad (45)$$

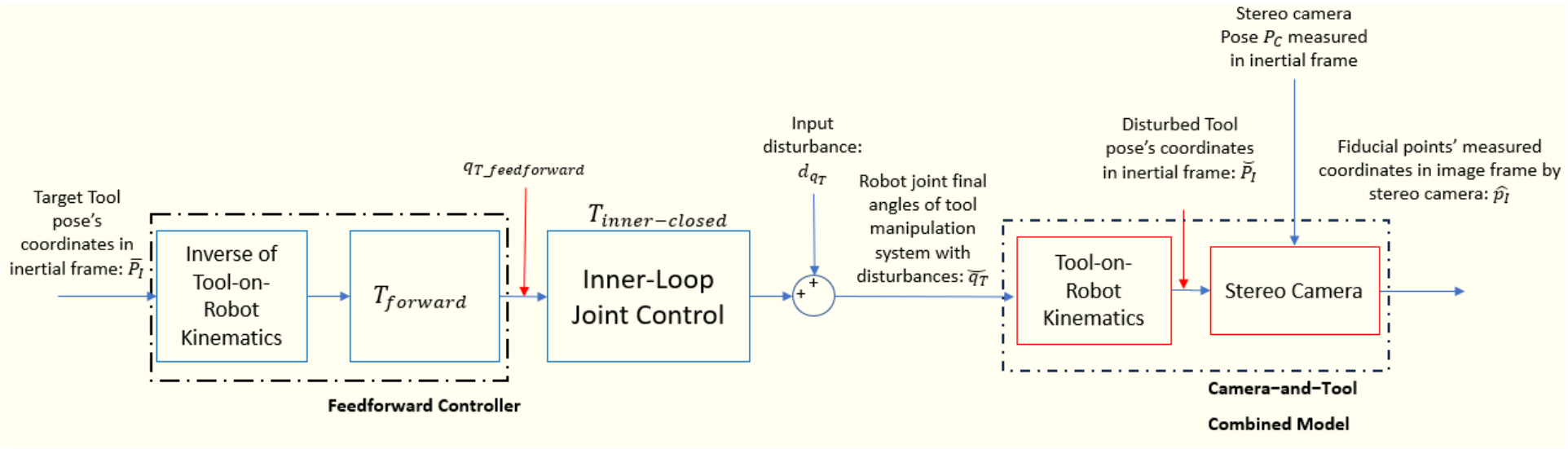

Figure 9. Block diagram of feedforward control design

### 4.4 Feature Estimation

As the camera is static (not moving with the robot arm), the tool pose cannot be recognized and measured visually if it is outside the camera range of view. To tackle this problem, the 2D feature (image coordinates of the tool points) can be estimated from the same combined model in Equation (30) with the joint angles $q_T$ as input shown in Figure 3:

$$\widetilde{p_I} = \mathcal{F}(q_T) \tag{46}$$

## 5. Simulation Results

### 5.1 Response of tool's pose in the control system

In this section, we will simulate the closed adaptive control loop with different scenarios to check the performance and the stability of the proposed controller. For all scenarios, the tool is guided to a particular target position measured in the base inertial frame {O}: $\overline{P_I} = [-1m, 0.2m, 0.3m]^T$, where $m$ represents meters and the tool should be aligned vertically so that its rotational matrix equals to $[\bar{n}, \bar{s}, \bar{a}] = \begin{bmatrix} 0 & -1 & 0 \\ -1 & 0 & 0 \\ 0 & 0 & -1 \end{bmatrix}$, where $\bar{n}, \bar{s}, \bar{a}$ represent respectively the end-effector's directional unit vector of the yaw, pitch and roll in the inertial frame {O}. All scenarios are simulated with bandwidth of the inner-loop as $100 rad/s$ and the bandwidth of the outer-loop as $10 rad/s$.

In the following three plots, the blue lines represent the responses of the end-effector position, and the red lines are the targets. In the trajectory plots, the circles represent the starting position of the end-effector, and the stars represent the target position of the end-effector. Black arrows are orientation direction of trajectory points that change over time, and red arrows are the target direction of the end-effector.

Three simulation scenarios are illustrated in Figure 10, Figure 11, and Figure 12, respectively. In the first, the tool (end-effector) starts at the pose: $P_I^o = [1.404m, 0.228m, 1.171m]^T$, and $[n_0, s_0, a_0] = \begin{bmatrix} -0.4893 & -0.0262 & 0.8717 \\ 0.2427 & 0.9560 & 0.1650 \\ -0.8377 & 0.2932 & -0.4614 \end{bmatrix}$, where two fiducial points are in side view of the camera. Therefore, their coordinates can be detected and measured during the entire control period. In the second and third scenario, the tool (end-effector) starts at the pose: $P_I^o = [1.285m, 0m, 1.57m]^T$, and $[n_0, s_0, a_0] = \begin{bmatrix} 0 & 0 & 1 \\ 0 & 1 & 0 \\ -1 & 0 & 0 \end{bmatrix}$, where two fiducial points are not inside view of the camera at beginning. Therefore, their coordinates must be estimated at the initial stage of the control loop. Disturbances of joint angles are added to the first and the third scenarios, as $d_q = [0.1°, 0.5°, 0.2°, 0.3°, -0.1°, 0.3°]$ to test the controller robustness against these input disturbances.

The response of Figure 10 and Figure 12 illustrate that the designed feedforward-feedback adaptive controller can reach the target in a stable manner even in the present of input disturbances. Compare orientation trajectories between Figure 11 and Figure 12, disturbances add instability in the transient response, but the control system is able to stabilize it in the steady state response. When the tool's coordinates are measurable by the camera (as shown in Figure 10), the feedforward and feedback controller work together to guide the tool to its target pose fast and precisely, which is indicated as a response within 0.3 seconds and small overshoots in transient response. However, when the tool is outside the view of the camera (as shown in Figure 11 and 12), the controlled system still converges fast (less than a second) but generates large overshoots in some of the position responses. The large overshoots may come from the accumulated disturbances that cannot be eliminated by the feedforward control without the intervention of the feedback control. The issue of the overshoots will be investigated in the future research and can be dealt with either by upgrading the camera with a wider range of view or by the using a switching algorithm, such as switching from a feedforward to a feedback controller, rather than the use of a continuous feedforward-feedback controller.

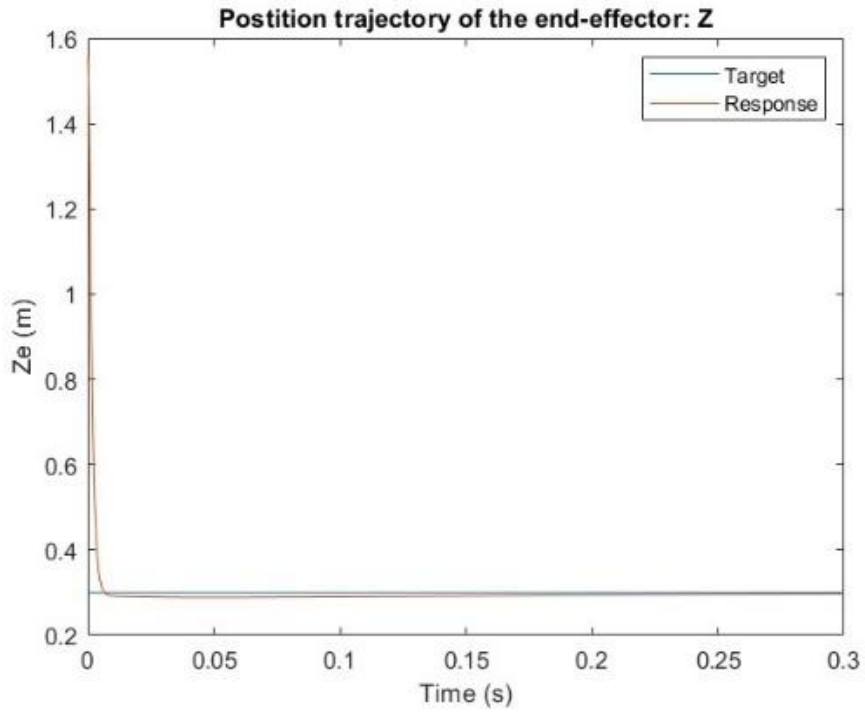

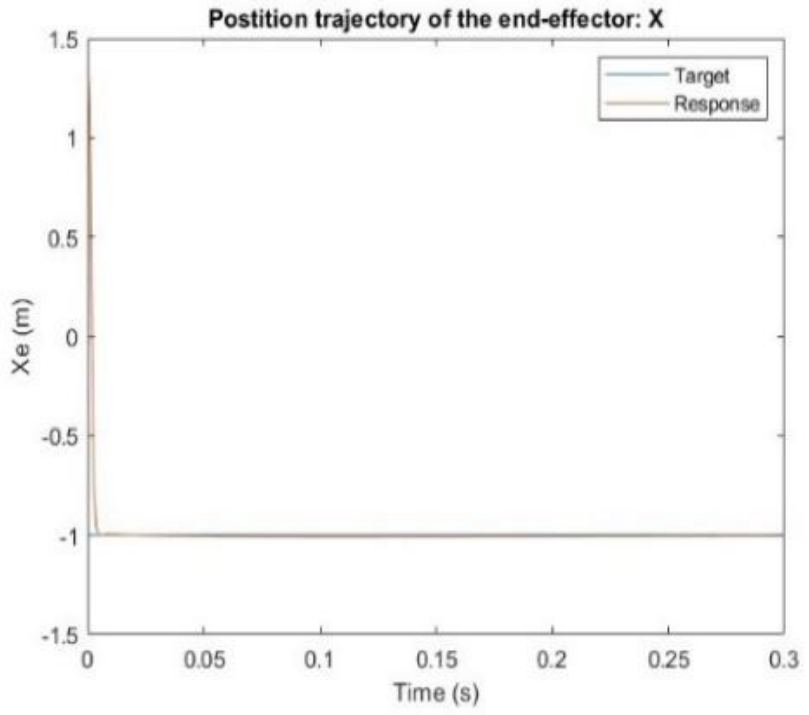

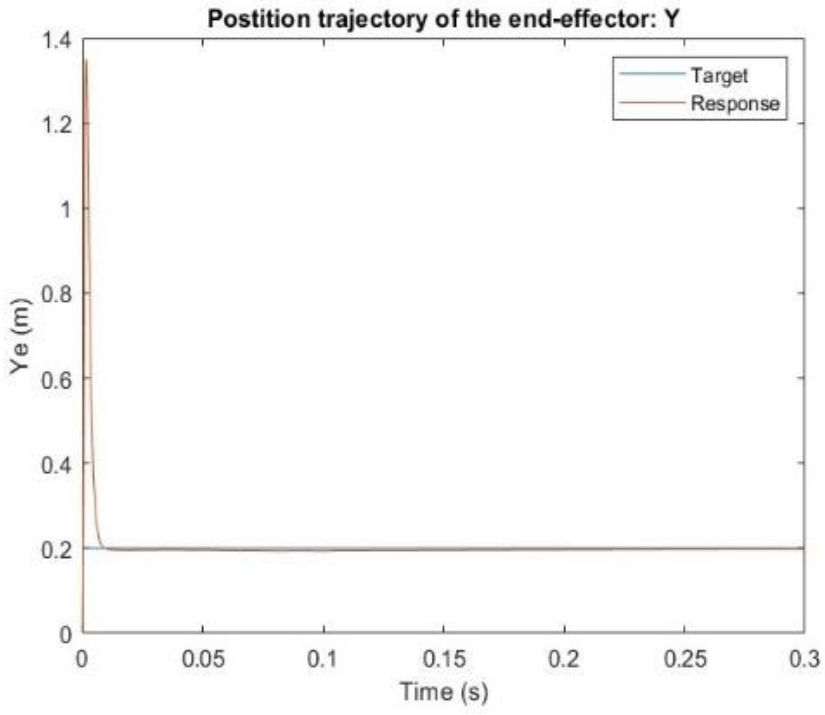

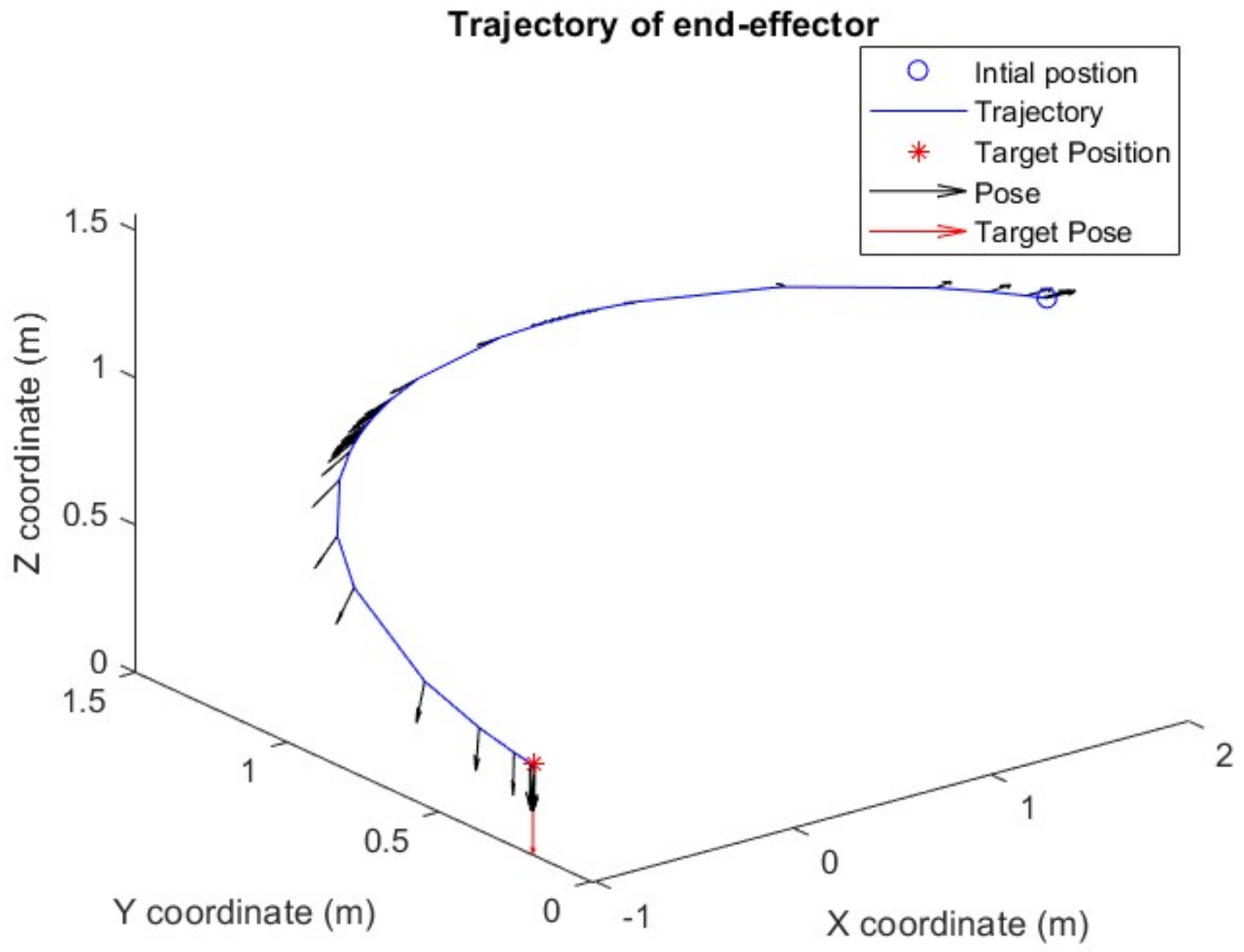

Figure 10. Response of the tool's (end-effector's) pose in scenario 1: Start in field view of camera with disturbances.

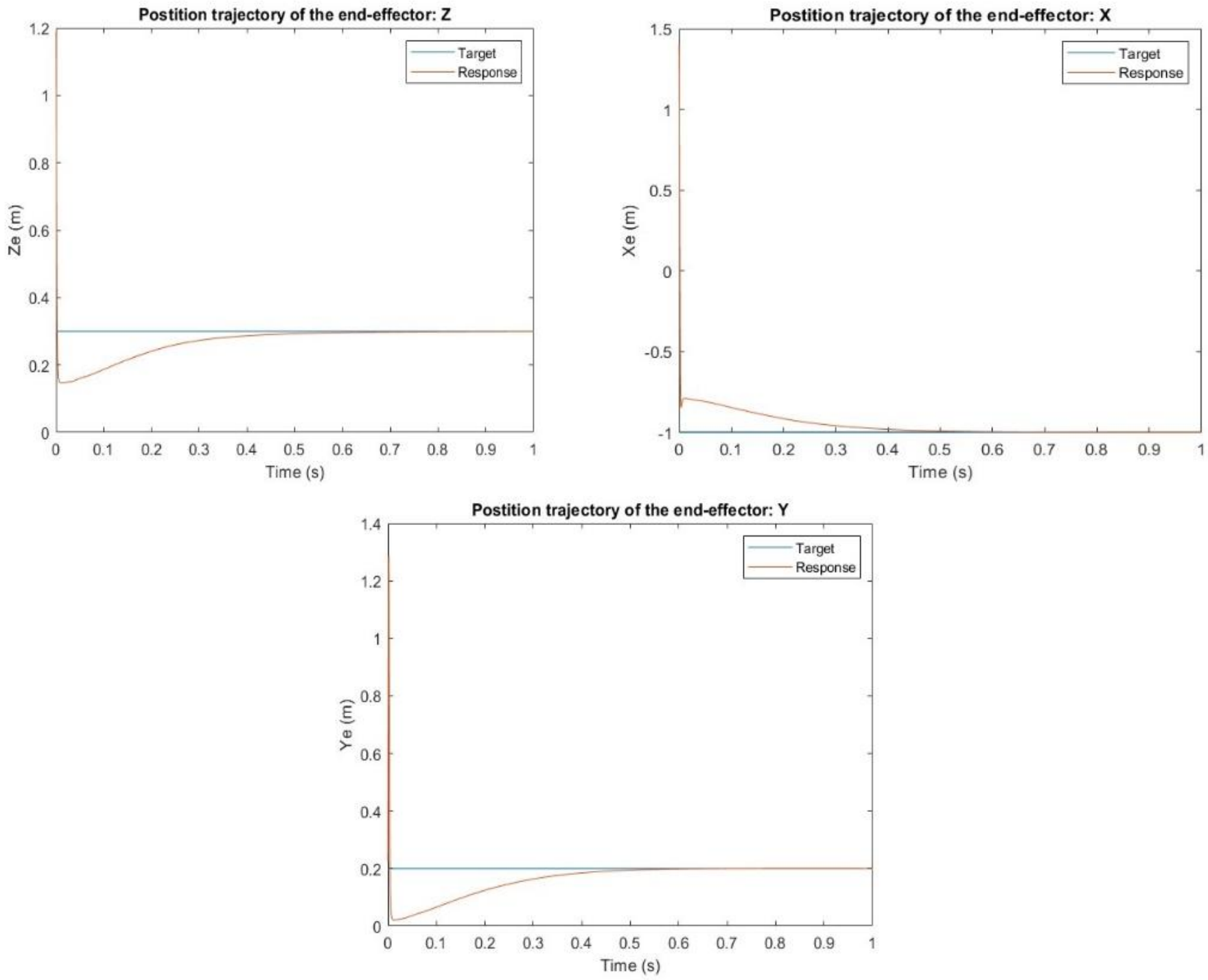

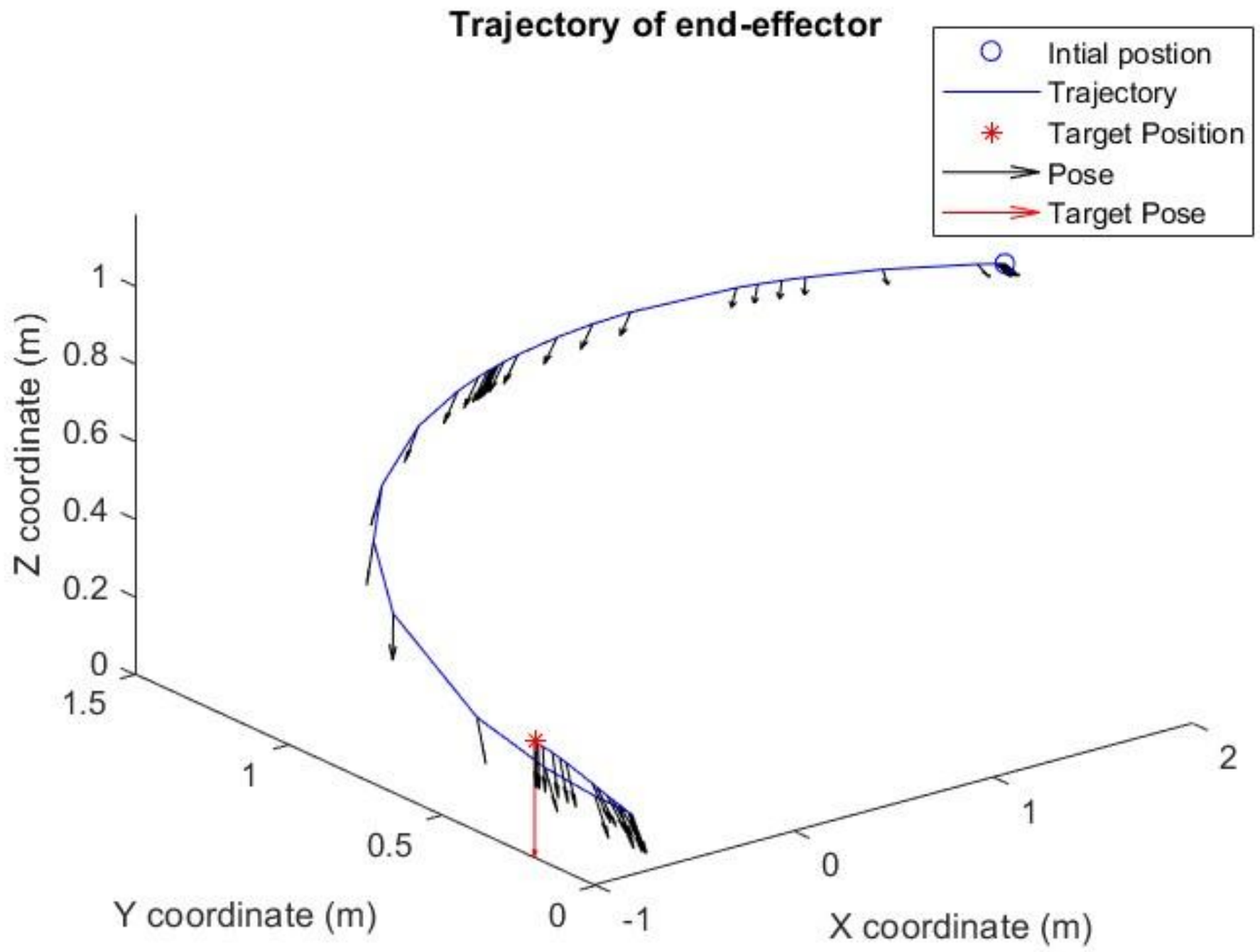

Figure 11. Response of the tool's (end-effector's) pose in scenario 2: Start outside field view of camera with no disturbances.

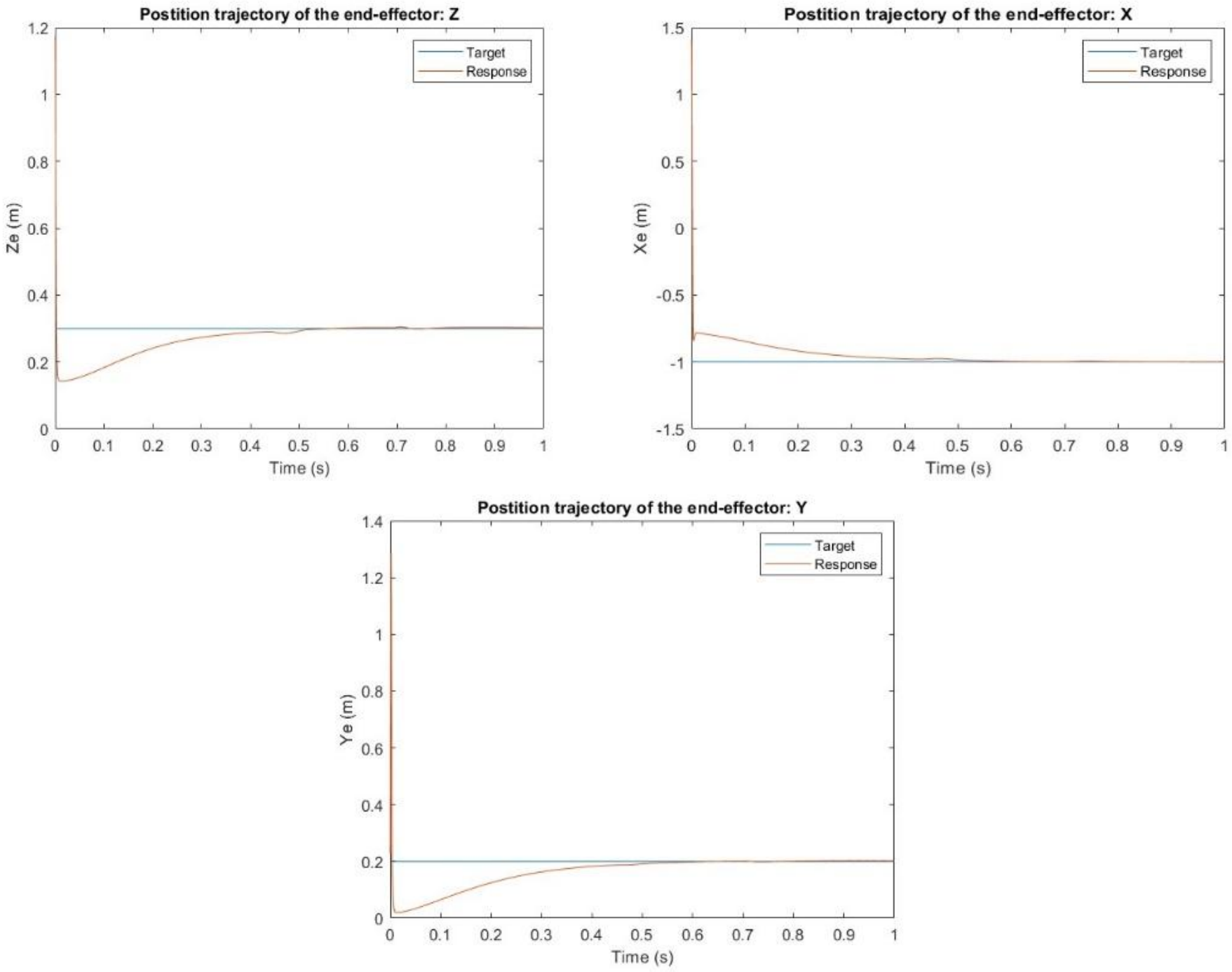

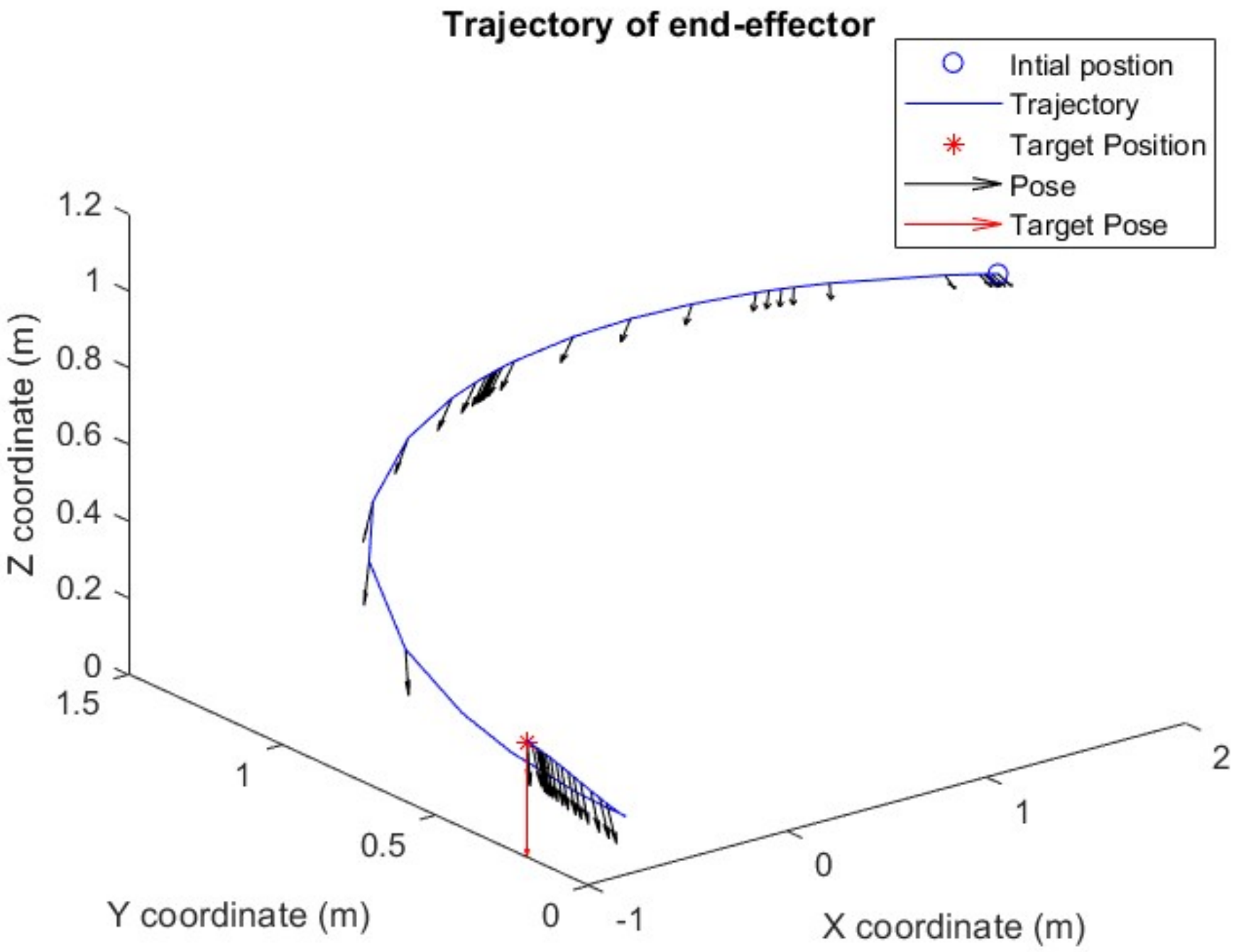

Figure 12. Response of the tool's (end-effector's) pose in scenario 3: Start outside field view of camera with disturbances.

### 5.2 Robustness analysis of Model uncertainties

In this section, we study the impact of varying some parameters on the steady state error of the end-effector X-coordinate position. The length of link 2 and link 4 of the robot manipulator $L_2$ and $L_4$ are chosen to vary as uncertainties because they have the two largest numerical dimensions among all geometric parameters of the robot manipulator.

Figure 13 shows the robustness test taken in simulation scenario 1. All points in the variation range have a steady state error which is less than 1%. We can use this as a standard to show the robust performance of the controller against model variation. From the plot, we can summarize that the designed control system is robust to model variation of the two parameters varying individually between 50% to 110%.

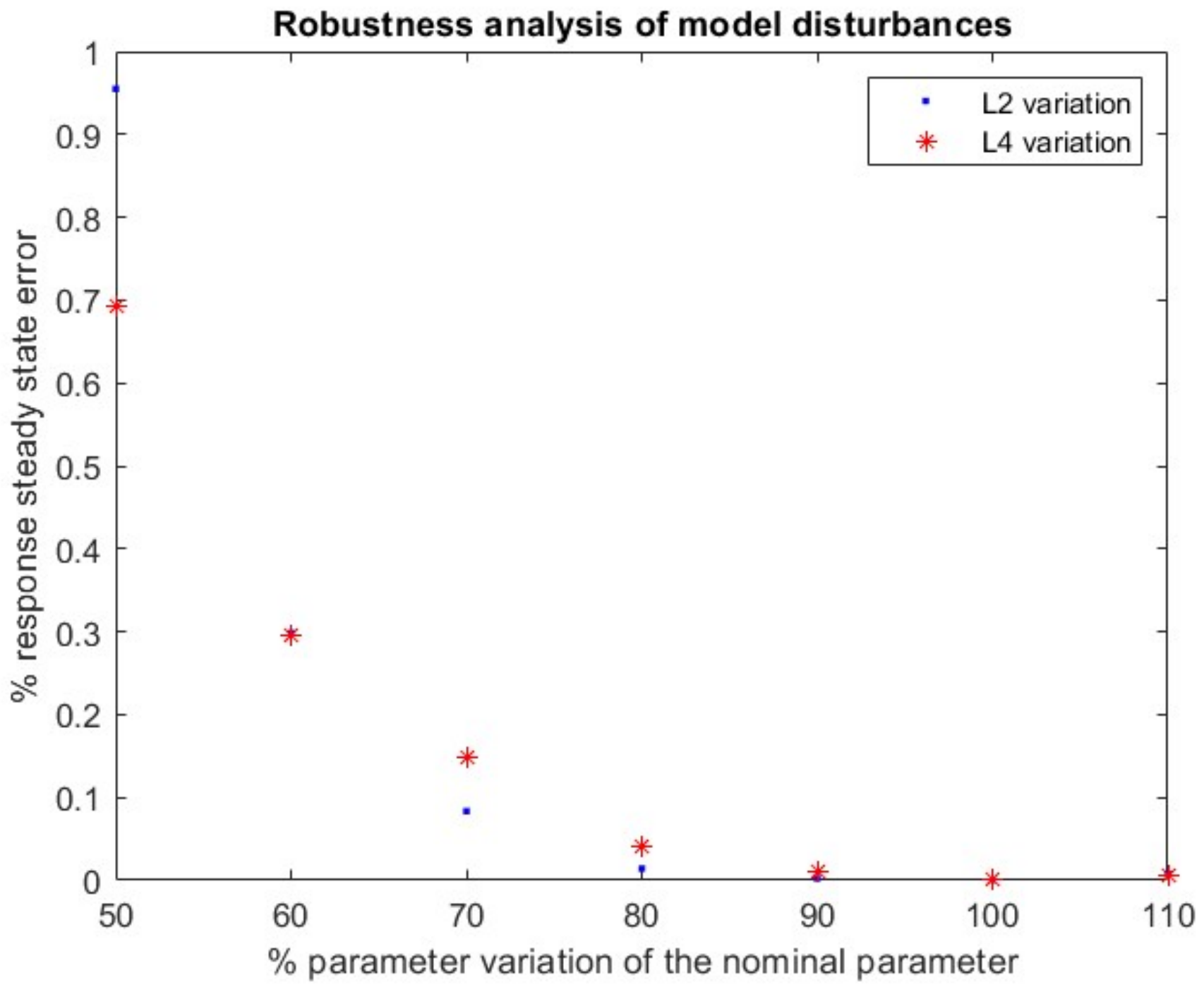

Figure 13. Robustness analysis of model uncertainties.

## 6. Conclusion

This paper proposes a new feedforward-feedback adaptive control algorithm based on Image Based Visual Servoing technique. The core of this method is to design a feedforward controller and feature estimation loop so that the tool's pose can be stabilized when it is outside the view of the camera. Furthermore, an adaptive Youla parameterization-based controller is developed in the feedback loop to ensure the stability and performance for the entire range of operations of the robot manipulator in the workspace. Various scenarios of the fastening and unfastening alignment problem have been simulated and robustness against input disturbances and model uncertainties have been examined. In summary, our control design has shown fast and robust performance in different scenarios so that it has a great potential for implementation on the real-world applications in automated manufacturing fields.

## 7. Appendix

Table 2. Parameter values of ABB IRB 4600 Robot Manipulator [31] and Zed 2 Stereo camera [33]

| **Parameters (Robot manipulator)** | **Value** |
|---|---|
| Length of Link 1: $L_1$ | 0.495 m |
| Length of Link 2: $L_2$ | 0.9 m |
| Length of Link 3: $L_3$ | 0.175 m |
| Length of Link 4: $L_4$ | 0.96 m |
| Length of Link 1 offset: $a_1$ | 0.175 m |
| Spherical wrist: $L_t$ | 0.135 m |
| Tool length of screwdriver: | 0.127 m |
| **Parameters (Zed 2 Stereo Camera)** | **Value** |
| Focus length: $f$ | 2.8 mm |
| Baseline: $b$ | 120 mm |
| Angle of view in width: $\alpha$ | 86.09° |
| Angle of view in height: $\beta$ | 55.35° |